\documentclass[letterpaper, 10 pt, conference]{ieeeconf}  

\IEEEoverridecommandlockouts                              

\usepackage{cite}
\usepackage{amsmath,amssymb,amsfonts}
\usepackage{algorithmic}
\usepackage{subfigure}
\usepackage{graphicx}
\usepackage{textcomp}
\usepackage{float}
\usepackage{lipsum}  
\usepackage[labelfont=bf]{caption}

\usepackage{todonotes} 

\title{\LARGE \bf
Autonomous Overhead Powerline Recharging \\ for Uninterrupted Drone Operations
}

\author{Viet Duong Hoang$^{*1}$, \textit{IEEE Member}, Frederik Falk Nyboe$^{*1}$, \textit{IEEE Member}, \\ Nicolaj Haarhøj Malle$^{*1}$, \textit{IEEE Member}, Emad Ebeid $^{1}$, \IEEEmembership{IEEE Senior Member}
\thanks{$^{*}$ These authors equally contributed to this paper}
\thanks{$^{1}$ Department of Mechanical and Electrical Engineering (DME), University of Southern Denmark, Odense, Denmark
        {\tt\small vdh|ffn|nhma|esme@sdu.dk}}%
}

\begin{document}

\maketitle
\thispagestyle{empty}
\pagestyle{empty}

\begin{abstract}
    We present a fully autonomous self-recharging drone system capable of long-duration sustained operations near powerlines. The drone is equipped with a robust onboard perception and navigation system that enables it to locate powerlines and approach them for landing. A passively actuated gripping mechanism grasps the powerline cable during landing after which a control circuit regulates the magnetic field inside a split-core current transformer to provide sufficient holding force as well as battery recharging.

    The system is evaluated in an active outdoor three-phase powerline environment. We demonstrate multiple contiguous hours of fully autonomous uninterrupted drone operations composed of several cycles of flying, landing, recharging, and takeoff, validating the capability of extended, essentially unlimited, operational endurance.


\end{abstract}

\vspace{0.25cm}
For an online video demonstration, please visit: \\ https://www.youtube.com/watch?v=C-uekD6VTIQ

\section{Introduction}
\label{sec:introduction}


The deployment of multi-rotor drones for corridor inspection of linear assets, such as overhead power lines, has revolutionized the domain of infrastructure inspection. These missions have proven to be indispensable for acquiring accurate and real-time data concerning the structural integrity of linear assets, contributing significantly to their maintenance and safety. Nevertheless, the inherent constraint of limited flight time due to drone battery limitations remains an obstacle to realizing the full potential of these missions. This paper addresses this limitation by demonstrating a novel strategy for during-mission drone recharging, thereby promoting complete system autonomy and extending mission duration indefinitely.

\begin{figure}[t]
    \centering
    \includegraphics[trim={0 10 0 0},clip, width=0.95\linewidth]{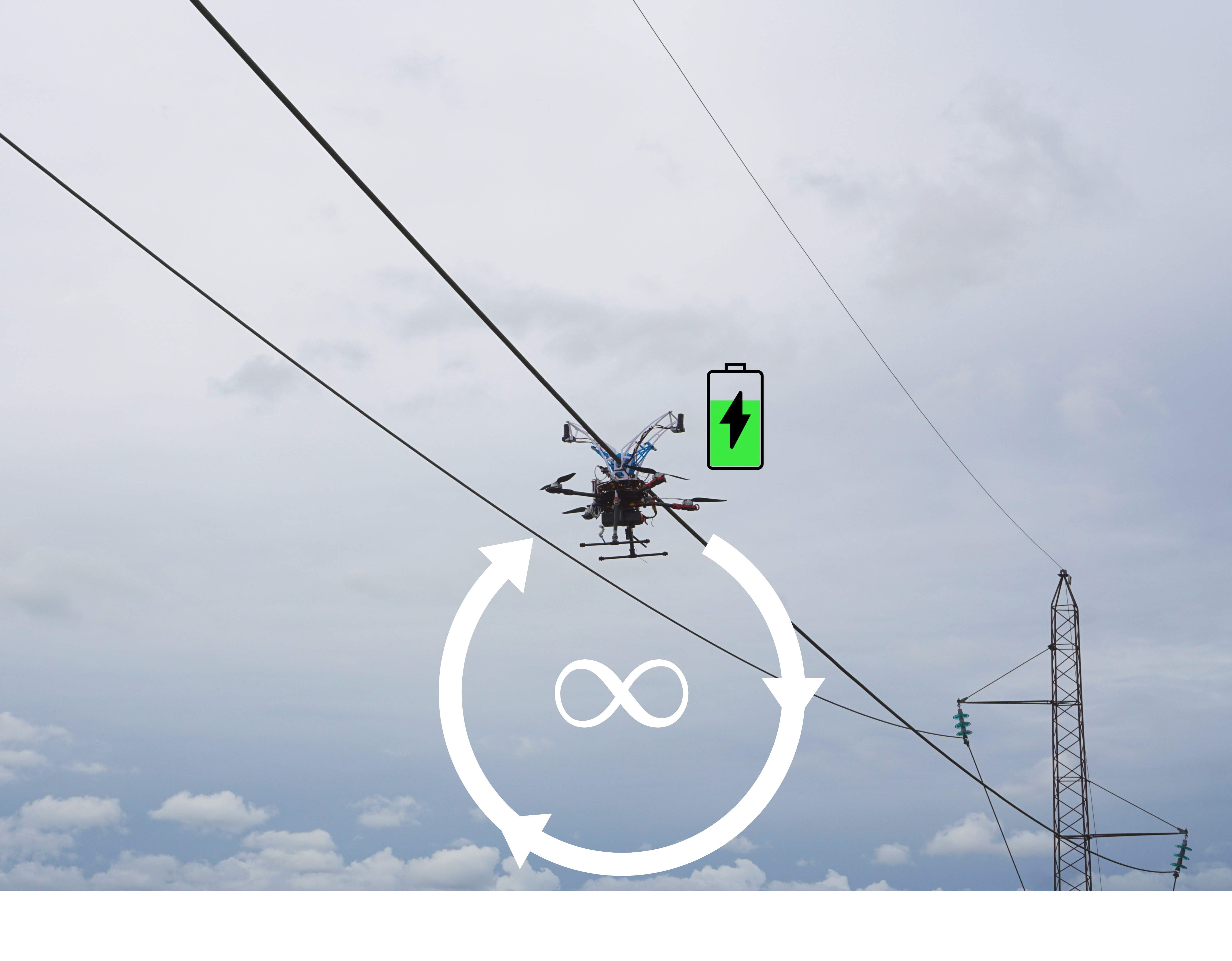}
    \vspace{-8pt}
    \caption{Autonomous drone recharging from powerline.}
    \label{fig:title}
    \vspace{-0.5cm}
\end{figure}

The work described in this article extends our previous work in drones for powerline operations; it integrates previously proposed systems comprising a powerline perception system \cite{Malle2022}, a drone low-level autonomy and powerline navigation system \cite{Nyboe2023}, a powerline energy harvesting system \cite{viet2023Advanced, viet2024Manipulating}, and mitigation of electromagnetic interference on the drone in \cite{skriver2022experimental}. On top of this, we propose in this article a novel gripper design as well as a mission autonomy system and demonstrate the system's capacity to execute continuous flight near powerlines while recharging on demand. The contributions of this work are
\begin{itemize}
    \item a novel gripper design to grasp the powerline that minimizes the force needed for closure;
    \item a magnetic manipulating circuit to maintain the holding force and harvest energy based on the powerline current level and battery state;
    \item a mission autonomy system guiding the drone to cycle between inspection and recharging; and
    \item to the best of our knowledge, a first-in-the-world system with the ability to sustain operation throughout many inspection/charging cycles powered by energy harvesting from powerlines in a real outdoor environment.
\end{itemize}

\begin{figure}[t]
    \centering
    \vspace{6pt}
    \includegraphics[width=0.95\linewidth]{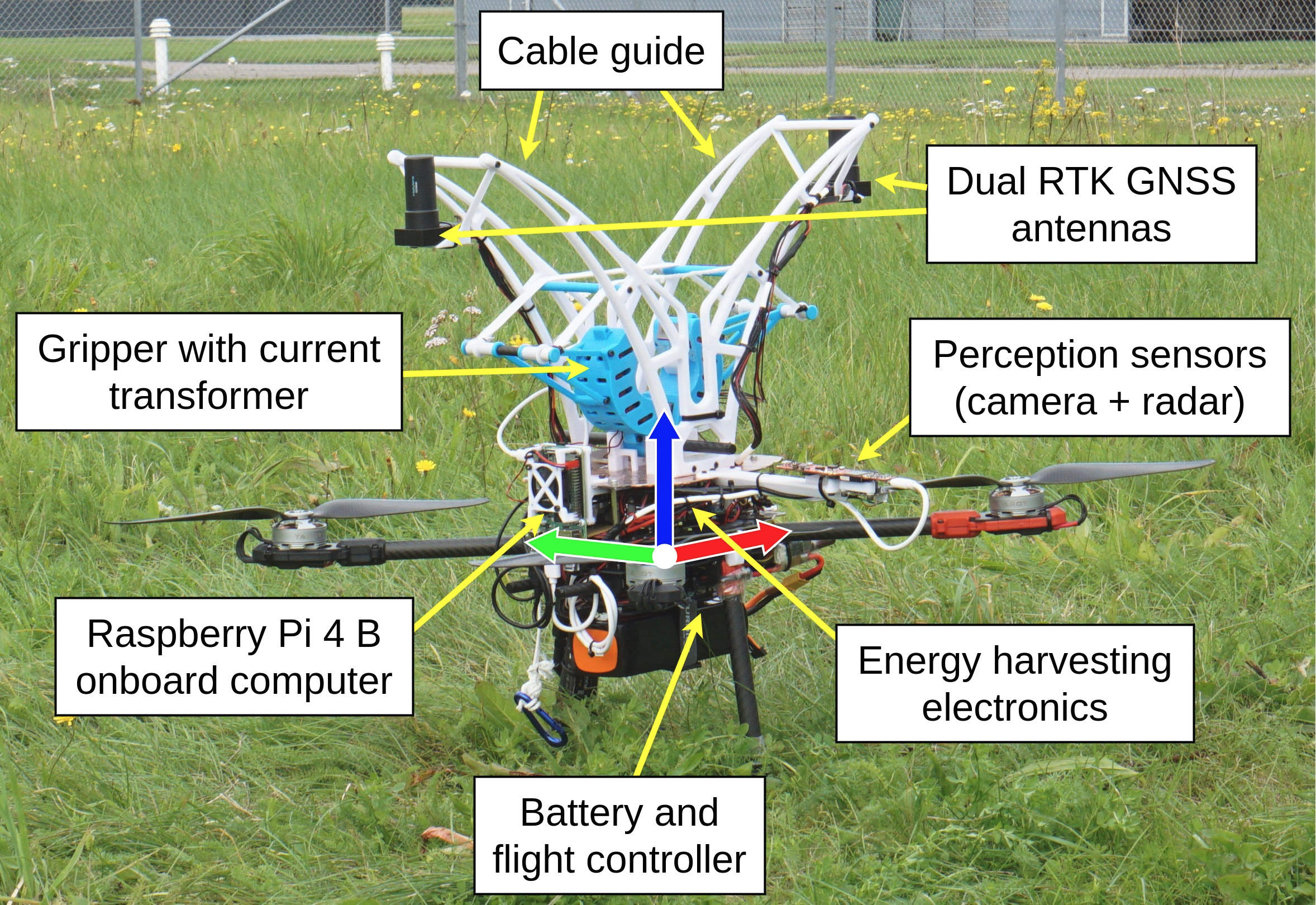}
    \caption{Components of the self-charging drone.}
    \label{fig:hw_diagram}
    \vspace{-0.4cm}
\end{figure}

The system, seen charging in Fig. \ref{fig:title}, is detailed with its main components in Fig.~\ref{fig:hw_diagram}. The base vehicle frame is a Tarot~650 Sport~\cite{tarot} equipped with a generic quadcopter propulsion system consisting of 33~cm carbon fiber propellers, 6S 380~KV BLDC motors, 30~A ESCs, and a 7000~mAh 30C six-cell LiPo battery. A CUAV Pixhawk V6X Autopilot (V6X)~\cite{v6x} controls low-level flight while a connected Raspberry Pi 4 B (RPi4B)~\cite{rpi4b} performs onboard computing of perception and autonomy algorithms. The onboard perception sensors include an IWR6843AOPEVM mmWave radar device~\cite{mmwave} and a generic global shutter USB camera. The top of the drone is outfitted with a large powerline cable guide on either side of which an RTK GNSS antenna is mounted. The cable guide supports the powerline landing maneuvers while the dual GNSS antennas provide a heading source that does not rely on magnetometers which may be compromised by the high electromagnetic interference from the live powerline. The gripping mechanism is mounted inside the cable guide and supports the drone's weight and recharges its battery while landed on a powerline. The gripper orientation relative to the flight controller means that the drone's y-axis is parallel to the powerline cable direction when performing landing, charging, and takeoff. Sandwiched between the gripper and the flight controller are the magnetic control and charging circuits which control the electromagnet of the gripper. The system has been designed to reduce weight where possible, and the overall system weighs 4.3 kg including battery.

\section{Related Work}
\label{sec:related_work}
Previous work has been done in solving the battery depletion problem for UAVs. In \cite{7822448}, Choi et al. presents the design of a autonomous system and wireless charging station, enabling drones to land and recharge. Similarly, Saviolo et al.~\cite{10161503} presented a system enabling in-flight recharging from a charging unit. However, both of these approaches require infrastructure, which might not be feasible to install in a remote environment.

In order to save energy while inspecting the powerline, some studies proposed the idea of landing on and rolling along the line \cite{8461250, hamelin2019discrete}. In those publications, the authors present a system which lands semi-autonomously on top of a powerline. The system does not function autonomously and independently, and the work does not address recharging. 

In other work, charging stations mounted on the cables or next to the towers have also been considered as a solution to recharge drones in the field, as it can achieve a high charging performance \cite{0b09b6dab0d34649bc6703b8722692c4, app131810175, obayashi2021400}. However, installing the charging stations around the transmission line infrastructure is costly and time-consuming, and significantly limits the flexibility of the technology to be used in remote areas. 

Drones with integrated energy harvesters and chargers are a more promising method that has been recently investigated in various work~\cite{stewart2022lightweight, kitchen2020design, drones5040108}, including some of the authors' previous work \cite{irosyee, viet2023Advanced, vom2020drones, 10161506}. However, none of the previously proposed systems demonstrate continuous operation. Additionally, increased performance has been achieved by the authors as demonstrated in this work.

\section{Gripper design}
\label{sec:gripper}

\begin{figure}[b]
    \centering
    \includegraphics[width=0.9\linewidth]{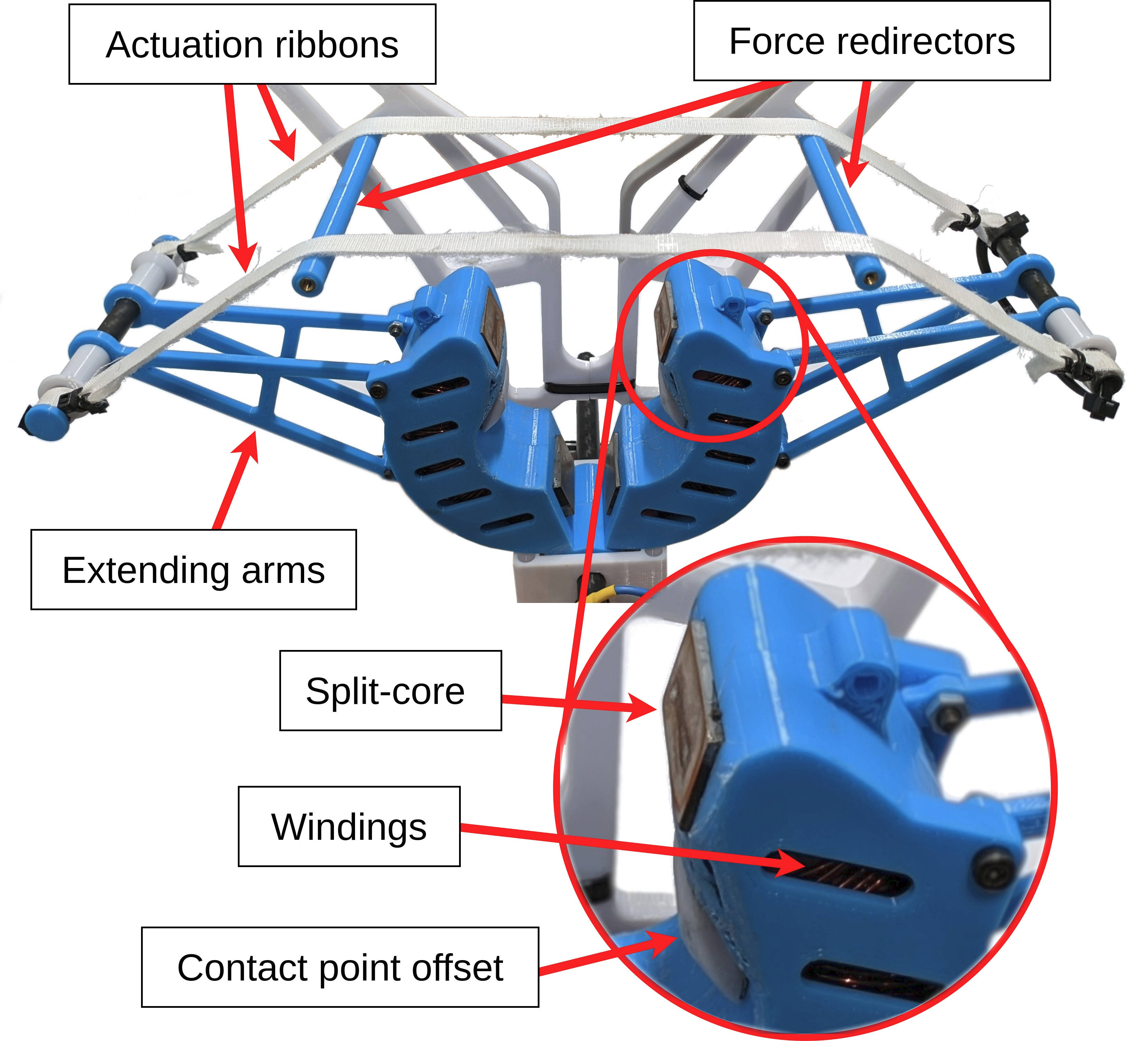}
    \caption{Components of the gripper mechanism.}
    \label{fig:grippermechanism}
    \vspace{-0.2cm}
\end{figure}

\begin{figure*}[h]
\centering
\begin{minipage}{.33\textwidth}
    \centering
    \includegraphics[height=4cm]{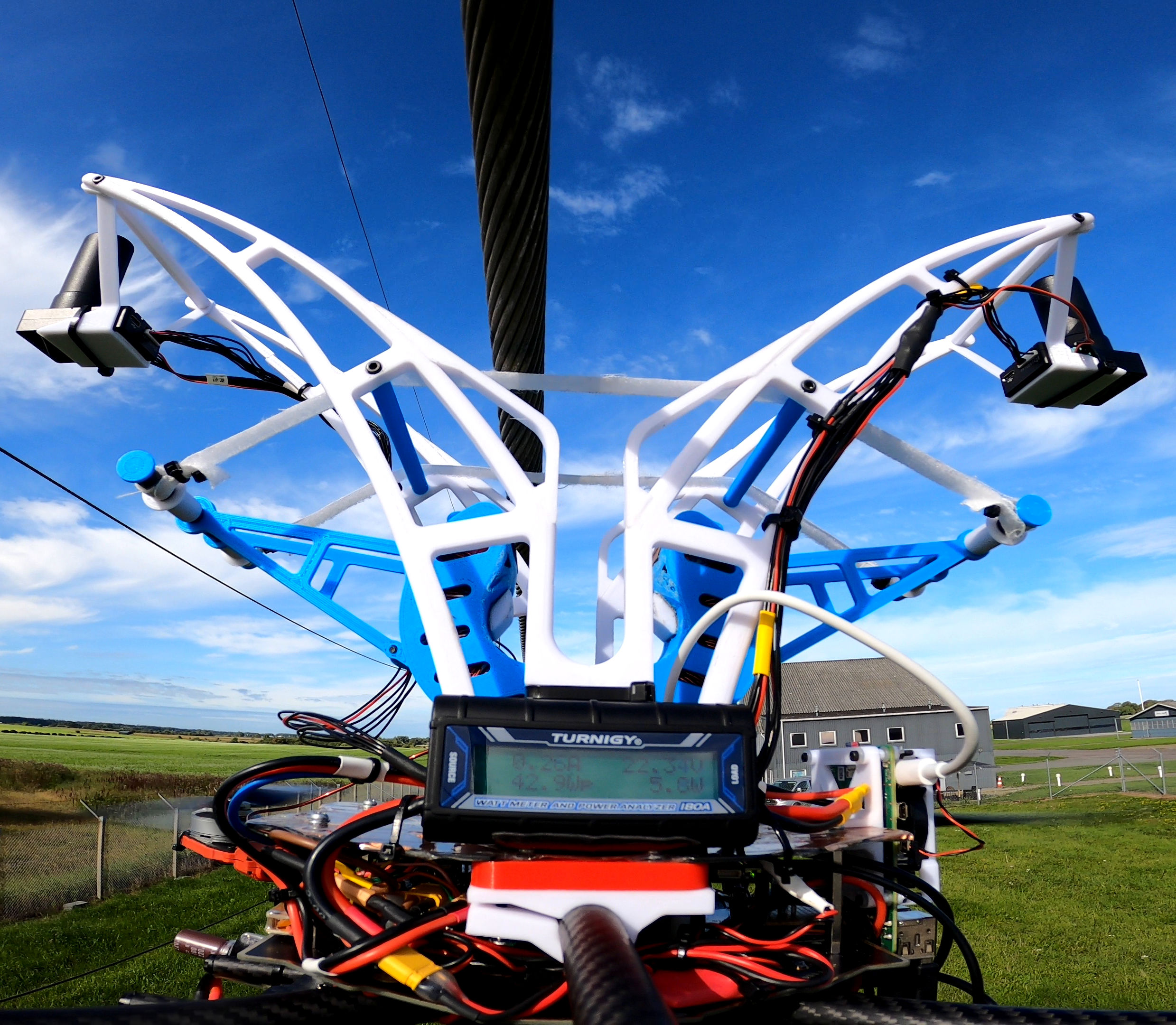}
    \captionof{figure}{Gripper in open state.}
    \label{fig:gripopen}
\end{minipage}%
\hfill
\begin{minipage}{.33\textwidth}
    \centering
    \includegraphics[height=4cm]{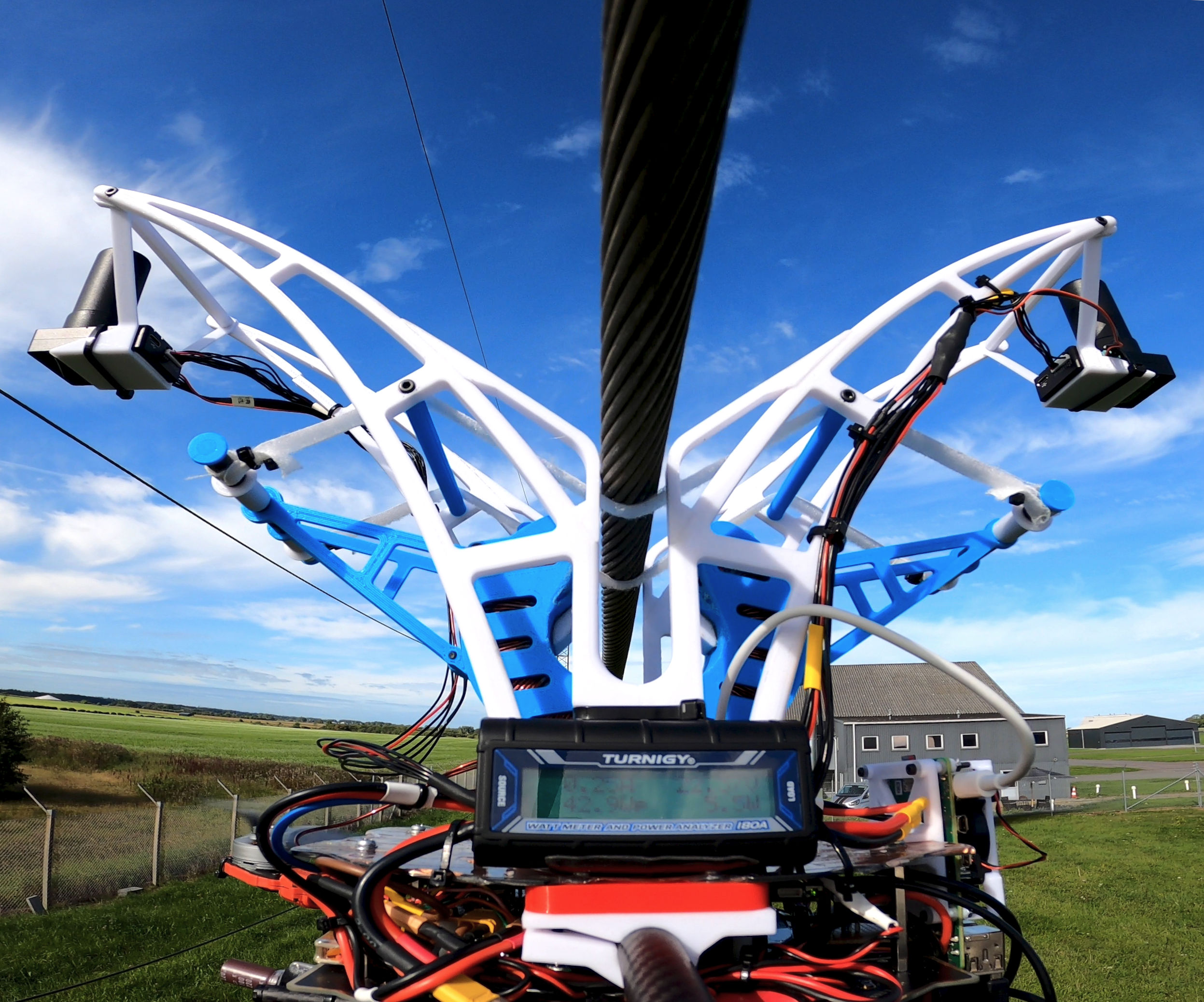}
    \captionof{figure}{Gripper engaged by powerline.}
    \label{fig:gripengaged}
\end{minipage}%
\hfill
\begin{minipage}{.33\textwidth}
    \centering
    \includegraphics[height=4cm]{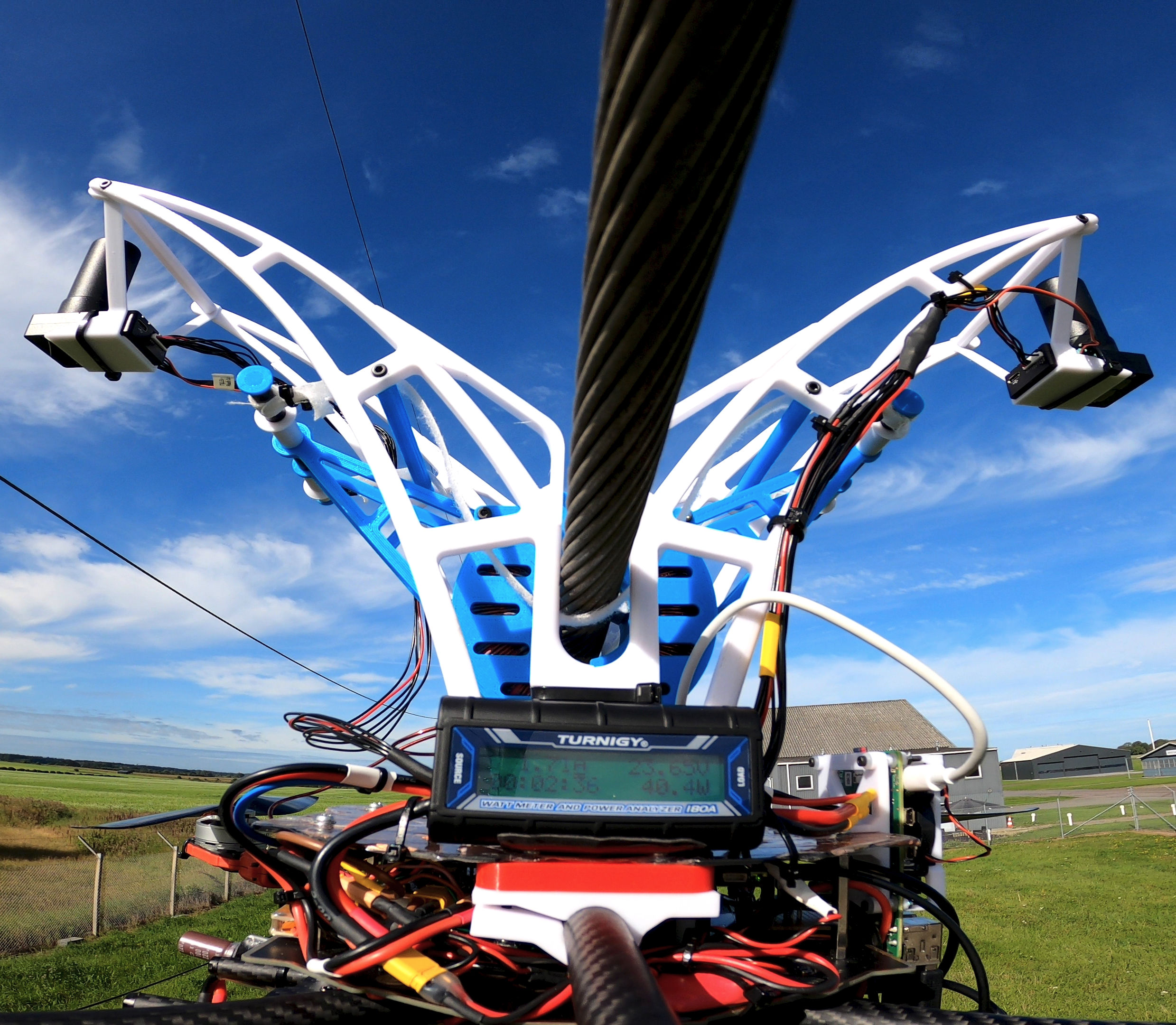}
    \captionof{figure}{Closed gripper.}
    \label{fig:gripclosed}
\end{minipage}
\vspace{-0.5cm}
\end{figure*}

The gripper design is a significantly upgraded version of the original design first presented in \cite{viet2023Adaptive}. The core principle is a split-core current transformer that is passively actuated when approaching the overhead powerline cable. As the drone captures the cable in the cable guide, the slope of the cable guide directs the drone's upward trajectory such that the open core of the gripper is guided toward the powerline cable. The powerline cable will contact the gripper actuation ribbon, seen in Fig. \ref{fig:grippermechanism}, approximately halfway down the cable guide. As the drone continues its upward trajectory, this will transition the gripper from the open state, as seen in Fig. \ref{fig:gripopen}, to the engaged and closed states (Fig. \ref{fig:gripengaged} and Fig. \ref{fig:gripclosed}). The actuation ribbon is connected to extending arms on both halves of the split-core case, and a downward force on the ribbon will force the two halves together. By routing the actuation ribbon over two carefully placed spacers inside the cable guide, the downward force on the ribbon is redirected to be approximately tangential to the rotational motion of the split-core halves, thereby significantly reducing the required actuation force. Rotation of the split-core halves occurs around a rigid carbon fiber tube that connects the split-core to the drone frame. 

The gripper mechanism includes several features to increase the chance of success when attempting to land on a powerline cable \cite{patent2023drone}. The 3D-printed split-core casing features mounting points for the extending arms as well as openings for ventilating the current transformer during operation. This is done to prevent overheating the windings when the magnetic control circuit is active. Another addition is the powerline cable contact point offset. This small component is added to move the cable contact point inside the core towards the center of the core which significantly increases holding force during powerline landings. The casing itself is also shaped in a way that allows the cable to enter the core even if there is a significant angle between the drone's roll angle (rotation around red arrow in Fig. \ref{fig:hw_diagram}) and the slope of the cable. This is achieved by removing material from potential impact and pinch points. The most prominent feature of the mechanism, the cable guides, also aim to relax the constraints on the autonomy system by increasing the area with which the cable can be caught. The tips of the guides are separated by 45 cm, allowing up to 22.5 cm of misalignment in the trajectory toward the cable. These measures make it possible for the system to function in adverse conditions, such as during windy weather or if the cable oscillates, which might not otherwise be possible.

The current design of the gripper requires less than 200 g ($\approx$2N) of downward force to completely close the gripper. This makes it suitable even for low thrust-to-weight ratio platforms that may otherwise lack the upward thrust to actuate a passive gripping mechanism. The passive actuation means that the gripper can also be used as a sensor to determine if the drone has successfully reached the cable, removing the need for other sensors. The closing time of the gripper entirely depends on the upward velocity of the drone and can therefore be fully closed in a fraction of a second. Once the gripper is closed, the magnetic control circuit is responsible for maintaining a sufficient holding force while simultaneously charging the drone's battery by manipulating the magnetic field inside the split core, as explained in Sec. \ref{sec:magnetic}.

\section{Magnetic control circuit}
\label{sec:magnetic}
\begin{figure}[b]
    \vspace{-8pt}
    \centering
    {\includegraphics[width=0.8\linewidth]{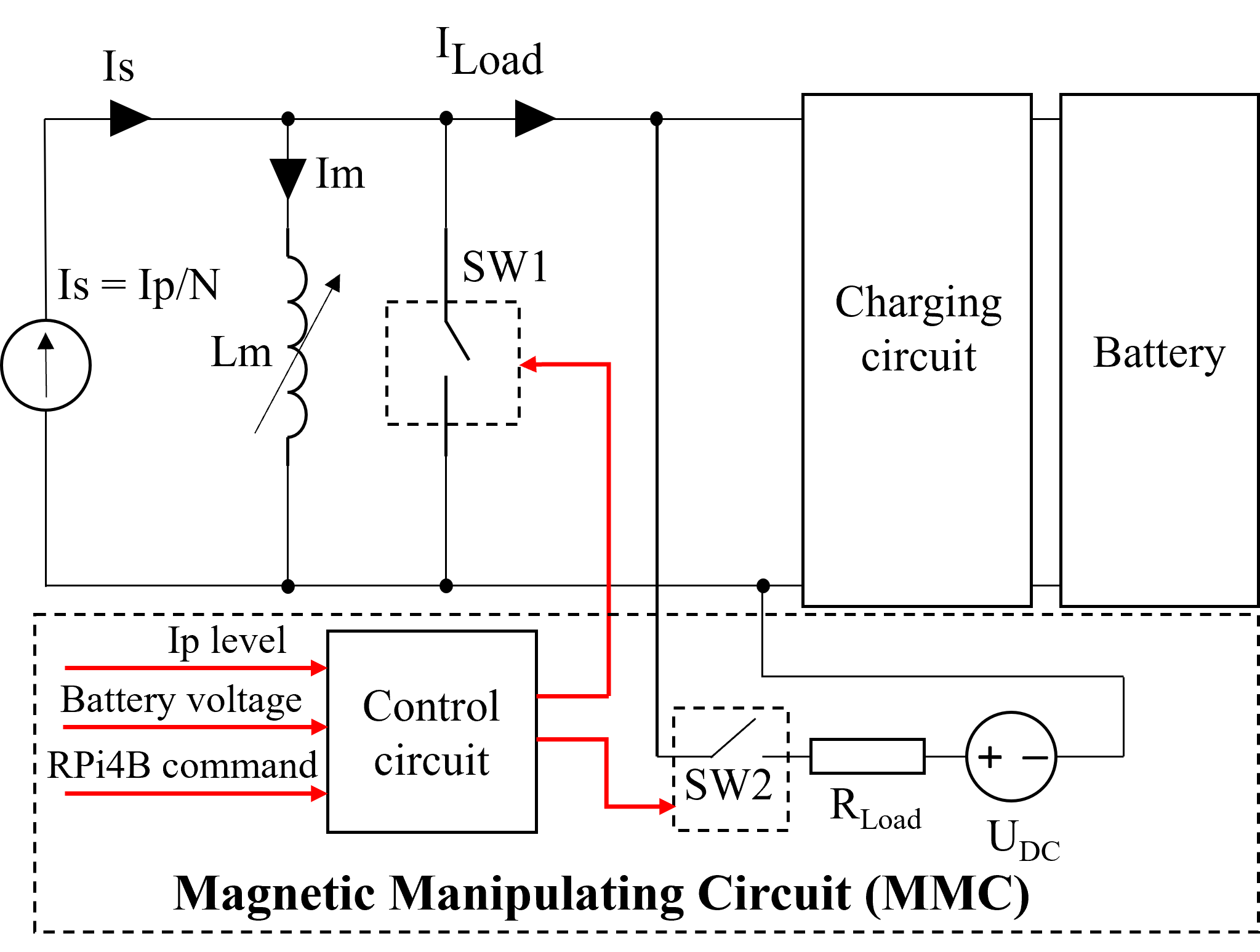}}
    \caption{Equivalent circuit of the current transformer with the MMC and the charging circuit}
   \label{fig:equivalent_circuit}
\end{figure}

The magnetic gripper is made of a current transformer that works as a gripper and an energy harvester. As the magnetic force is used to hold the drone on the line, it is essential to maintain this force under different powerline current conditions. The current $I_m$ in the equivalent circuit in Fig.~\ref{fig:equivalent_circuit} is the magnetizing current that magnetizes the inductive core. By controlling $I_m$, it is possible to control the holding force \cite{viet2023Adaptive, viet2024Manipulating}.

In the authors’ previous work \cite{viet2023Adaptive}, the magnetic gripper operated in two modes: DC mode (mode~1) and charging mode (mode~2). When the powerline current $I_p$ is low, the magnetic field is not strong enough to hold the drone. The DC current from the battery is used in this case, which is controlled by SW2. When the line current is high enough, and the battery voltage is low, the charging mode is activated. 

\begin{figure}[h!]
    \centering
    {\includegraphics[width=0.8\linewidth]{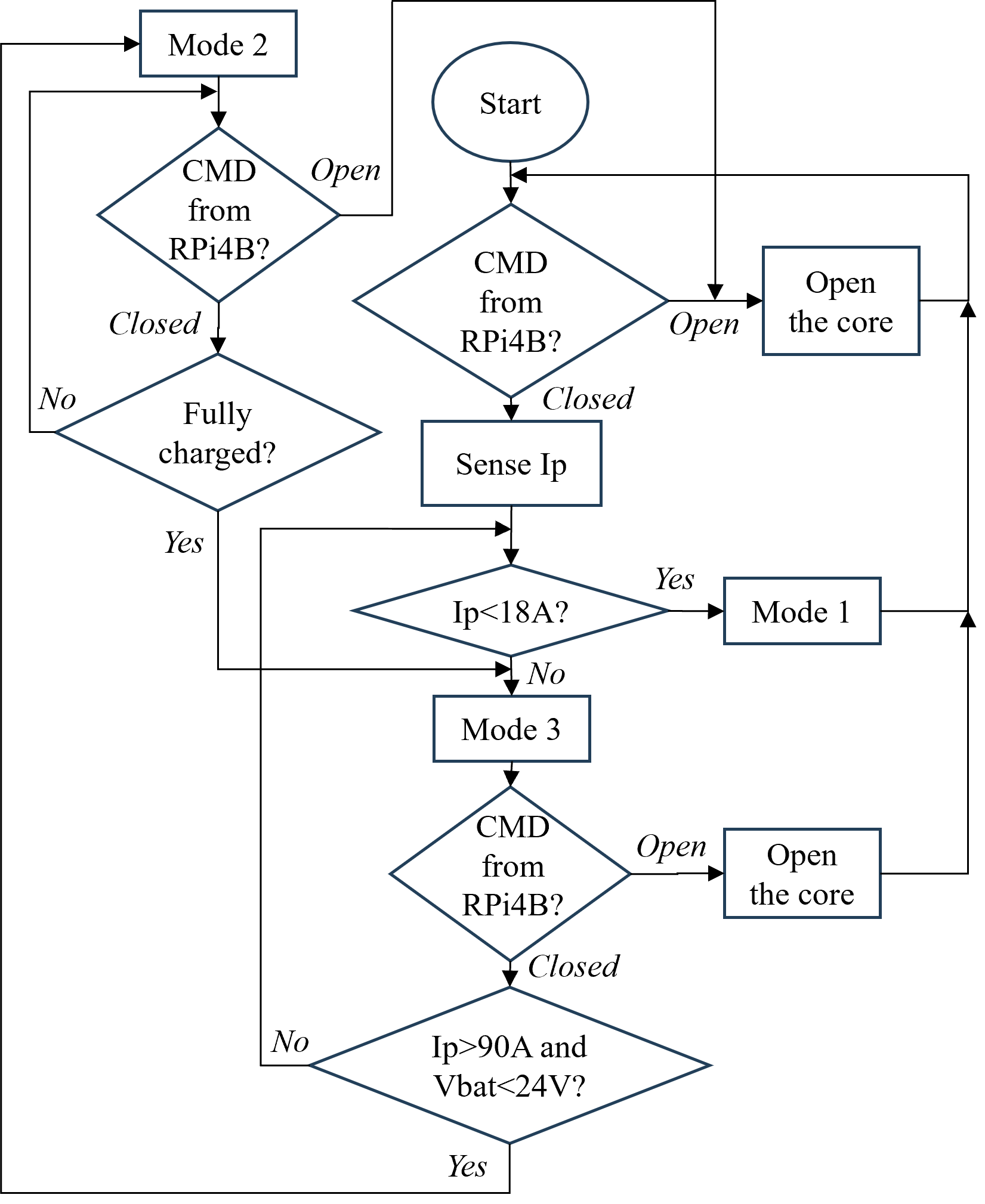}}
    \caption{MMC circuit algorithm to determine the operating modes based on the command from RPi4B, powerline current level and the battery state.}
   \label{fig:flow_chart}
   \vspace{-0.2cm}
\end{figure}

\begin{figure*}[h!]
    \centering
    \includegraphics[width=0.8\linewidth]{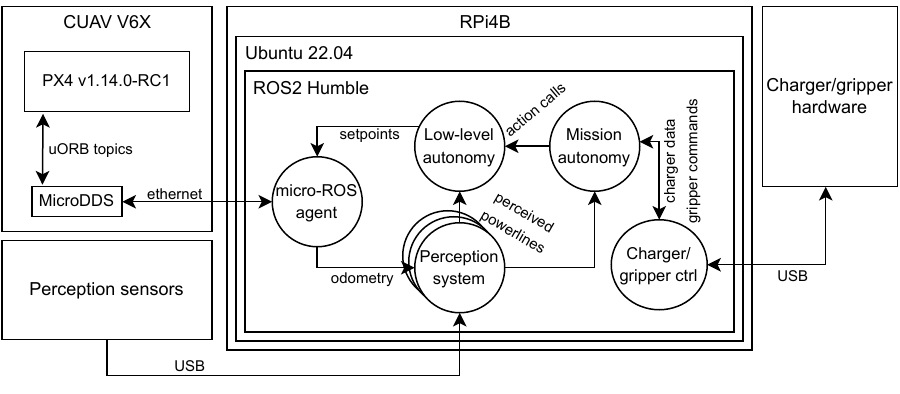}
    \vspace{-5pt}
    \caption{Computational system diagram showing the conections and data flow of the autonomous system.}
    \label{fig:computational_system}
    \vspace{-0.4cm}
\end{figure*}

Besides these two modes, when $I_p$ is still high and the battery is fully charged, it is possible to exploit the AC magnetic field to maintain the grip instead of using DC energy from the drone (mode~3) \cite{viet2024Manipulating}. As the characteristic of the inductor is to resist the change in current flowing through it, it is possible to keep the magnetizing current in the same direction regardless of the alternating AC magnetic field. In one line cycle (50 Hz for this paper), SW1 is open for a short time during the positive part of $I_p$ to store magnetic energy in $L_m$ and closed during the remaining time (including the negative half of $I_p$) to maintain the energy to keep $I_m$ at a positive level.

The simple method to open the magnetic gripper is to short-circuit the winding (close SW1) of the current transformer, after which $I_m$ gradually decreases until the core opens after several seconds. This requires more complicated control algorithms in the autonomous flight system. In order to open the gripper quickly, SW1 is closed at the zero point of $I_m$. 

The gripper status (open or closed) is very important for the autonomous flight system to determine when the drone can turn off the motors after landing on the line or when it should leave the cable during the taking-off phase. In mode~1, the powerline current is sensed by measuring $I_{Load}$. In mode~3, $I_p$ is checked by sensing the SW1 current during the closed time. If the powerline current is available, the gripper status is set to “Closed”. Otherwise, it returns “Open”. In mode~2 (charging mode), the charging power is checked instead, as the output current $I_{Load}$ is the charging current in this case. The charging power is proportional to the powerline current $I_p$ \cite{viet2023Advanced}.

The SW1 in Fig. \ref{fig:equivalent_circuit} is also controlled to enhance the harvested power using the Transfer Window Alignment Method (TWA) with Perturb and Observe (P\&O) algorithms \cite{viet2023Advanced}. The magnetizing current $I_m$ is controlled in different operating modes and purposes: keeping $I_m$ stable to maintain the grip, opening the core quickly, and enhancing the charging power. Thus, the circuit used to control $I_m$ is called Magnetic Manipulating Circuit (MMC). The MMC receives Open/Closed commands from the RPi4B and uses the level of $I_p$ and battery status to determine the operating modes, as shown in the flow chart in Fig. \ref{fig:flow_chart}. The MMC also transmits data to the RPi4B for use by the autonomy system: battery voltage, charging power/states, gripper state, and operating modes.

\section{Computational System}
\label{sec:autonomy}

The computational system comprises the perception system using mmWave radar and RGB camera and the low-level autonomy and navigation system, both of which were presented in previous work, as well as the mission orchestration system as detailed in this section, and a driver for the charger/gripper hardware. Additionally, the system integrates with the \texttt{PX4 Autopilot} software \cite{px4} \texttt{v1.14.0-rc1} running on the V6X equipped on the drone. Finally, the system is implemented using \texttt{ROS2 Humble} \cite{ros2} as transport layer middleware and framework, running on top of \texttt{Ubuntu Server 22.04.3 LTS} \cite{ubuntu22} on the RPi4B. The V6X is connected to the RPi4B using ethernet and integrates neatly with \texttt{ROS2} using \texttt{Micro XRCE-DDS} through the \texttt{Micro-ROS agent} \cite{microrosagent} running on the RPi4B side and \texttt{MicroDDS} running on the V6X side. A diagram of the computational system is seen in~Fig.~\ref{fig:computational_system}.

\subsection{Perception and Low-Level Autonomy}
The perception system fuses the mmWave radar data with the RGB camera images by extracting the powerline direction from the image and projecting the mmWave point cloud onto the vertical plane perpendicular to the powerline direction; each point is then tracked using a Kalman filter, incorporating odometry estimates from \texttt{PX4}, as explained in the authors' previous work~\cite{Malle2022}. This yields a representation where each line in the powerline setup is perceived as a point in cartesian space, along with the powerline direction. 

The low-level autonomy system utilizes the perception system output to expose a set of flight maneuver primitives aware of the perceived powerlines, such as \texttt{FlyToCable} and \texttt{LandOnCable}. The flight maneuver primitives are implemented as ROS2 actions and can thus be called from elsewhere in the system. Underneath, trajectories are continuously planned using linear Model Predictive Control (MPC) realizing the flight maneuver, and the setpoints are streamed to \texttt{PX4}, as was outlined in the authors' previous work~\cite{Nyboe2023}.

Together, the perception and low-level autonomy systems form the basis for programming of flight missions around powerlines at a high level of abstraction by providing information about the environment as well as providing a simple interface for execution of flight maneuver primitives.

\subsection{Charger/Gripper Control}
The charger/gripper control module constitutes a \texttt{ROS2} node which interfaces with the charger/gripper hardware through a serial connection over USB. The node receives data from the hardware about the current charger state, such as battery voltage, charging power, and charging mode, as well as information about the gripper, i.e. whether it is closed around a powerline or open. The node publishes this information as \texttt{ROS2} topics.

The node also handles commanding of the gripper to close and open over the same serial connection. The triggering of a gripper command happens upon a call to a \texttt{ROS2} service, which the node exposes to the additional system. In this way, the gripper can be controlled from high-level mission logic.

\begin{figure}[h]
    \vspace{-5pt}
    \centering
    \includegraphics[width=\linewidth]{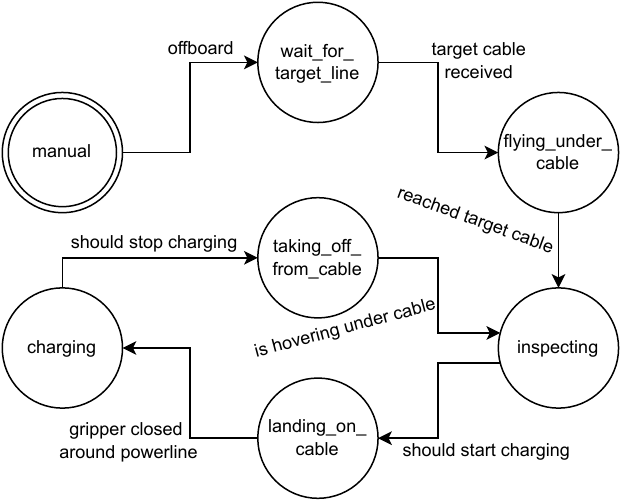}
    \vspace{-5pt}
    \caption{Mission autonomy state machine diagram.}
    \label{fig:mission_autonomy_fsm}
    \vspace{-0.4cm}
\end{figure}

\subsection{Mission Autonomy}
The mission autonomy is implemented for demonstration of the continuous operation capabilities of the system using the functionality exposed by the other system components. Utilizing a finite state machine, the mission toggles between states of inspecting and charging. Programming of advanced inspection behaviors is beyond the scope of this work, and as such the inspection state is simply represented by having the drone hovering under the cable. This state could be expanded to encompass actual inspection operations. The state machine diagram is seen in Fig.~\ref{fig:mission_autonomy_fsm}.

The conditions determining when the drone should start and stop charging are simply based on the battery voltage and can be configured with a low and high threshold. However, for practical reasons, \texttt{ROS2} services were implemented which could be called from the ground control computer to initiate or interrupt charging on command, thereby increasing the flexibility of the testing scenario.

During the \texttt{landing\_on\_cable} state, the node orchestrates the landing on the cable by calls to the low-level autonomy flight maneuver primitives, while issuing a command to close the gripper for the charger/gripper node. Finally, the drone is disarmed on the cable by commanding \texttt{PX4} to enter its land-mode. Should the cable landing be aborted by the low-level autonomy system due to violation of safety margins, the landing is reattempted. On the contrary, while in the \texttt{taking\_off\_from\_cable} state, the mission autonomy node arms the drone, opens the gripper, and leaves the cable, again using low-level primitives. 

This realizes a small, self-contained autonomous system, which demonstrates the continuous operation capabilities of the proposed technology.

\section{Experiments}
\label{sec:experiments}
The system is tested to verify its capacity for continuous operation in a real environment.

\begin{figure}[b]
    \vspace{-8pt}
    \centering
    \includegraphics[width=0.8\linewidth]{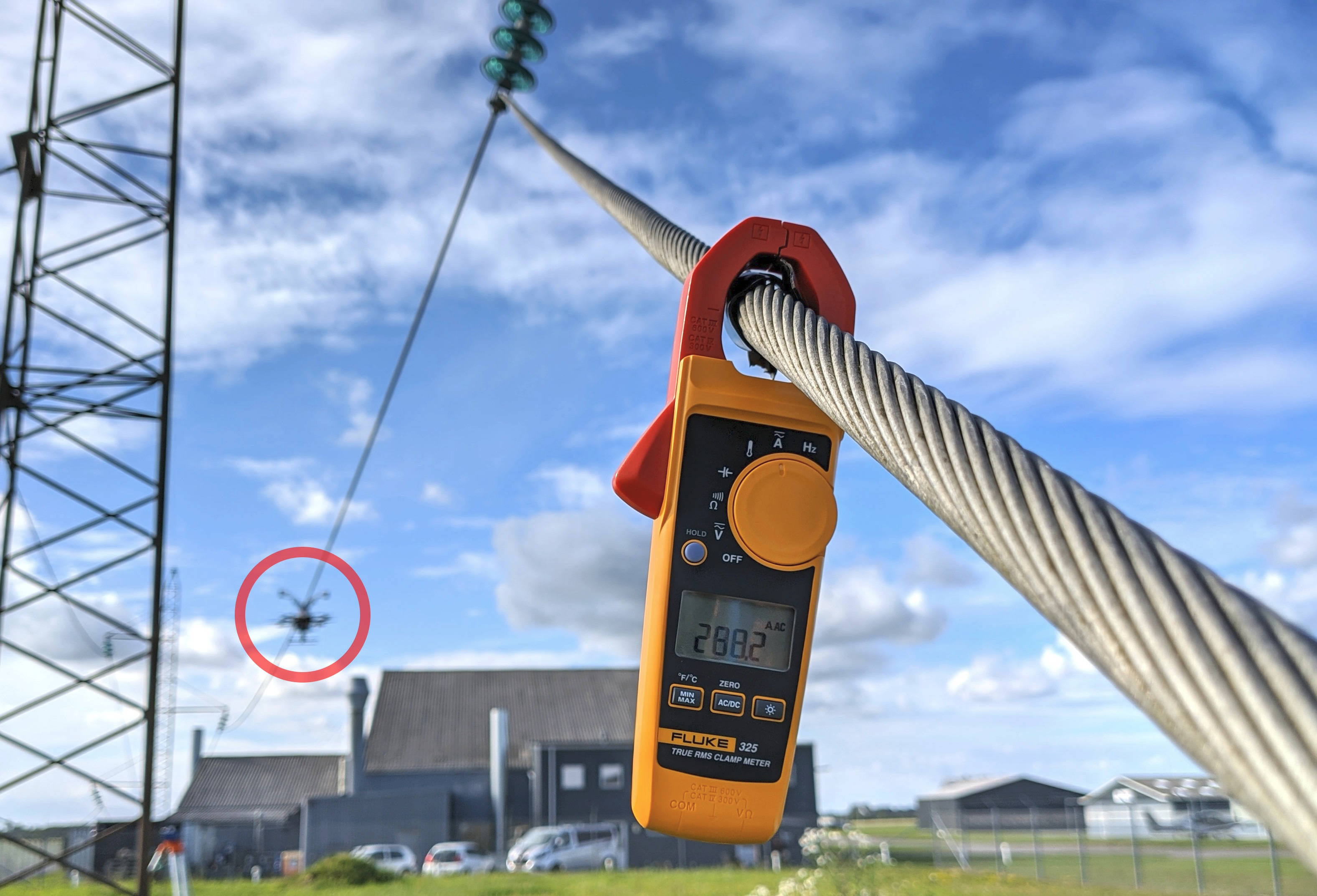}
    \caption{Powerline 50~Hz 288~A AC current during testing.}
    \label{fig:plcurrent}
\end{figure}

\subsection{Experimental Setup}
Testing was conducted at the outdoor three-phase powerline setup at HCA Airport in Odense, Denmark at the disposal of the authors, with each cable conducting upwards of 300~A. While the test voltage is low, we have previously shown how to make a drone resilient to upwards of 400 kV \cite{skriver2022experimental}. An example measurement of the cable current is seen in Fig. \ref{fig:plcurrent}. During the test, we had to comply with a legislative requirement of not having the drone in flight at the same time as manned flights in the area.

The integrated system was operated via a WiFi connection to a laptop running dedicated GUI software. From here, charging could be initiated and interrupted by the click of a button. Thus, the automatic starting and stopping of charging based on battery voltage levels could not be tested naturally. The initiation and interruption of charging was the only human intervention during testing.

Similarly, in order to test as many inspection/charging cycles as possible and thus limit charging time, charging was initiated and interrupted such that the battery was only discharged/charged by approximately 20\% each cycle. Data was captured using \texttt{ROS2} bag recordings.

\subsection{Results}

\begin{figure}[h]
    \centering
    \includegraphics[width=0.8\linewidth]{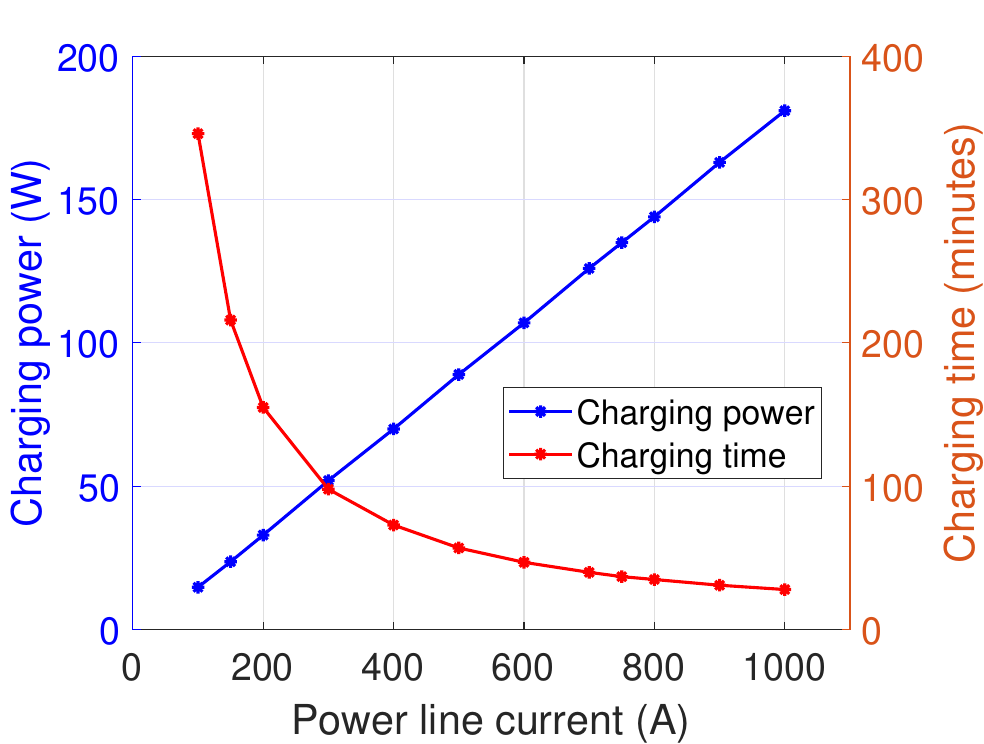}
    \caption{The relationship between the power line current level and charging power/charging time with 4.3 kg drone (including the 800 g current transformer) and 7 Ah 6S battery}
    \label{fig:ChargePowerAndTime}
    \vspace{-15pt}
\end{figure}

The 4.3 kg drone with components described in Section \ref{sec:introduction} is able stay airborne for 7.5 minutes. When the battery level drops below 45\%, the reduced voltage leads to insufficient thrust from the motors, making it difficult for the drone to achieve lift-off. Therefore, the charging time is only calculated for charging 55\% of the battery capacity. Fig.\ref{fig:ChargePowerAndTime} shows the relationship between the power line current and charging power/charging time. The charging power was tested in the range 100 A - 300 A, and the rest was calculated based on the linear characteristic. The charging power strongly depends on the power line's current level, managing only 15 W at 100 A but reaching 181 W at 1000 A. The charging time reduces significantly from 346 minutes (5.8 hours) to only 28 minutes when the power line current is increased from 100 A to 1000 A.

\begin{figure}[h]
    \vspace{-5pt}
    \centering
    \includegraphics[width=\linewidth]{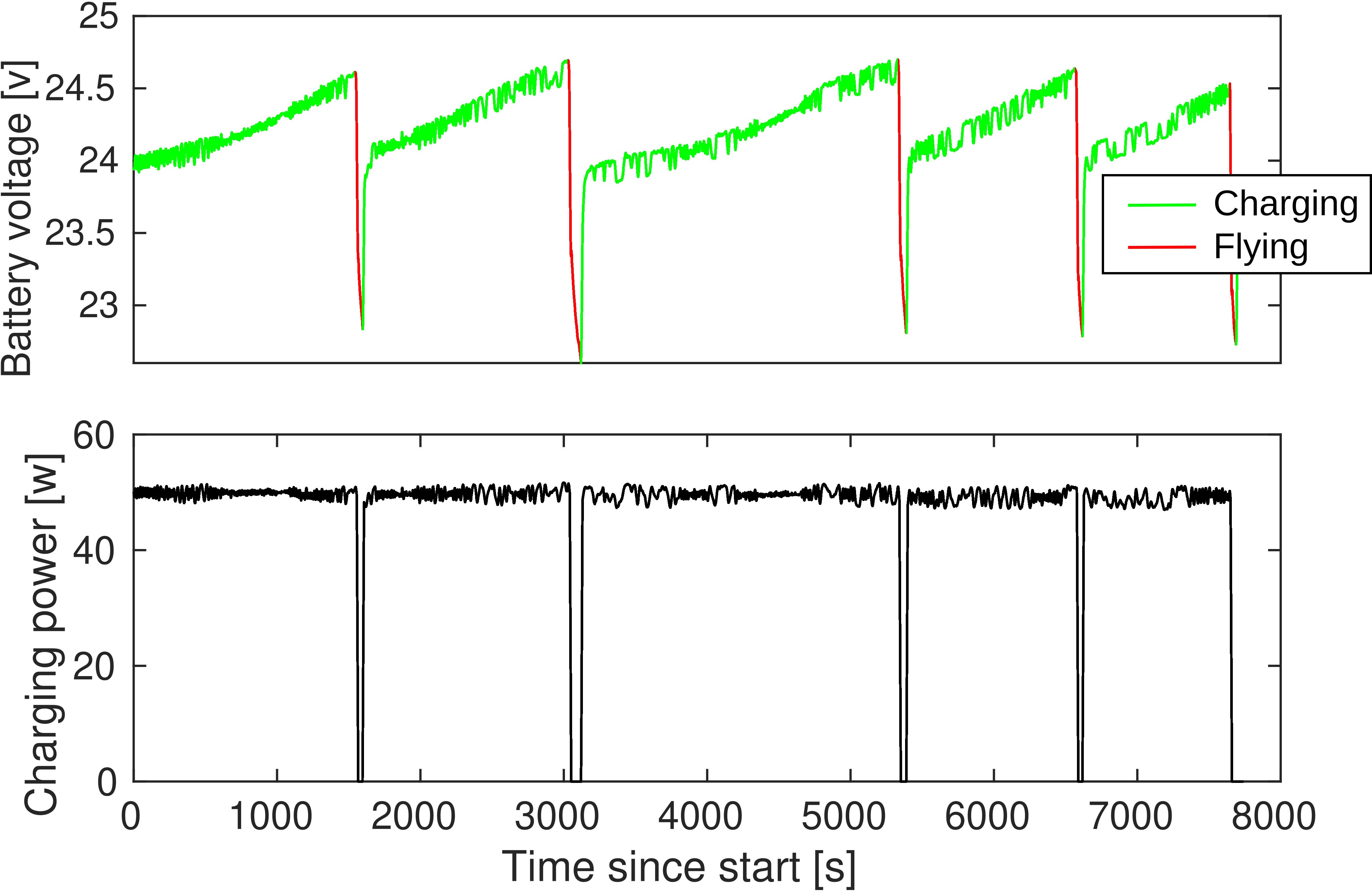}
    \vspace{-5pt}
    \caption{Battery voltage and charging power during the continuous inspection flight.}
    \label{fig:long_charging}
    \vspace{-0.4cm}
\end{figure}

The longest sustained test lasted more than two hours, comprising five inspection/charging cycles. 
\begin{figure}[!b]
    \centering
    \includegraphics[trim={0 0 0 0},clip, width=\linewidth]{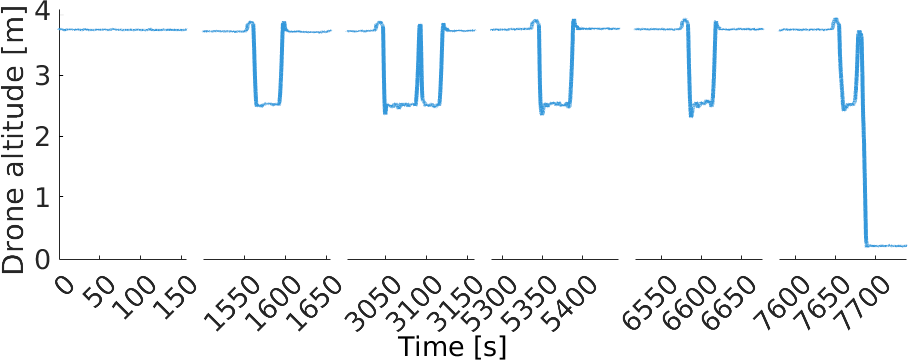}
    \caption{The drone altitude at six segments during the continuous mission.}
    \label{fig:altitude_plot}
    \vspace{-7pt}
\end{figure}
Fig.~\ref{fig:long_charging} shows the battery voltage and charging power over the course of the test, as measured in the charging electronics. It is seen that the voltage drops to less than 23~V during flight but is recovered fully during charging. Similarly, it is seen that 50~W charging power is achieved on the $\sim 300$~A line.

Fig.~\ref{fig:altitude_plot} shows the altitude of the drone, obtained from \texttt{PX4} odometry data, at six discontinuous segments during the test. It is seen how the altitude is constant while charging, and it displays the drone returning to the same altitude underneath the cable when in flight. During the second flight, it is seen that the landing is autonomously aborted and reattempted due to a violation of safety margins.

\begin{figure}[h]
    \centering
    \includegraphics[width=\linewidth]{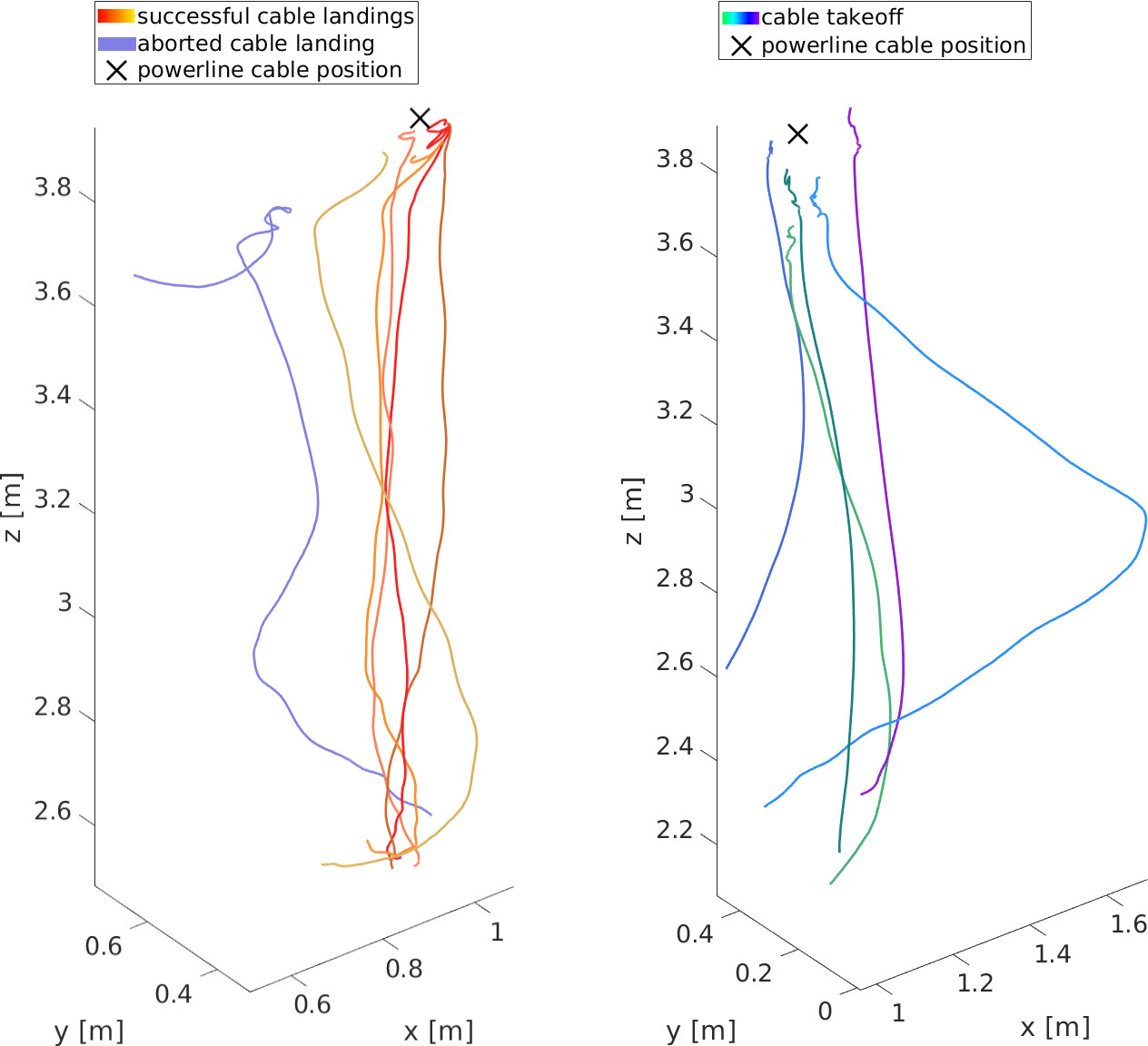}
    \caption{Trajectories of the drone during cable landing (left) and cable takeoff (right); axes units are in meters.}
    \label{fig:drone_traj}
\end{figure}

Fig.~\ref{fig:drone_traj} displays the drone trajectories during the performed cable landings and takeoffs, based on \texttt{PX4} odometry data, as well as the estimated cable position from perception system output. The single aborted landing is clearly visible.

\section{Conclusion}
\label{sec:conclusion}
In this paper, we demonstrated a fully autonomous drone system able to operate indefinitely by charging from powerlines on demand, enabling sustained inspection missions. A combined gripper and charger design was presented, upgrading the technology from our previous work. A mission autonomy system was presented, integrating previous modules for drone powerline operations. The integrated system was demonstrated to operate for more than two hours with five inspection/charging cycles, proving its feasibility.

Future work includes improving the system robustness and testing in more remote locations, expanding the complexity of the mission by adding inspection features, and investigating the system's robustness to adverse weather conditions.

\addtolength{\textheight}{-4.5cm}   





\section*{ACKNOWLEDGMENTS}
This project has received funding from the European Union’s Horizon 2020 research and innovation program under grant agreements No. 871479 (Aerial-Core) and No. 861111 (Drones4Safety) and Energifyn Fund.



\bibliographystyle{IEEEtran}
\bibliography{bibliography.bib}



\vspace{1.5cm}
Accepted for publication at IEEE International Conference on Robotics and Automation (ICRA) 2024.

© 2024 IEEE. Personal use of this material is permitted.  Permission from IEEE must be obtained for all other uses, in any current or future media, including reprinting/republishing this material for advertising or promotional purposes, creating new collective works, for resale or redistribution to servers or lists, or reuse of any copyrighted component of this work in other works.

\end{document}